\pdfoutput=1
\documentclass[11pt]{article}
\usepackage[preprint]{coling}
\usepackage{times}
\usepackage{latexsym}
\usepackage[T1]{fontenc}
\usepackage[utf8]{inputenc}
\usepackage{microtype}
\usepackage{inconsolata}
\usepackage{graphicx}

\usepackage{subcaption}
\usepackage{graphicx}

\usepackage{todonotes}
\usepackage{soul}           

\title{Disentanglement and Compositionality of Letter Identity and Letter Position in Variational Auto-Encoder Vision Models}

\author{
 \textbf{Bruno Bianchi\textsuperscript{1,@}},
 \textbf{Aakash Agrawal\textsuperscript{2}},
 \textbf{Stanislas Dehaene\textsuperscript{2,3,4}},
 \textbf{Emmanuel Chemla\textsuperscript{4,5}},
\\
 \textbf{Yair Lakretz\textsuperscript{4,5,@}},
\\
\\
 \textsuperscript{1}CONICET-UBA, FCEN, ICC-Departamento de Computación. Buenos Aires, Argentina.\\
 \textsuperscript{2}Cognitive Neuroimaging Unit, CEA, INSERM U 992, Université Paris-Saclay, NeuroSpin center, Gif/Yvette, France\\
 \textsuperscript{3}Collège de France, Paris, France\\
 \textsuperscript{4}PSL University, Paris, France
 \textsuperscript{5}LSCP, Ecole Normale Supérieure, CNRS, Paris, France
\\
 \small{
   \textbf{Correspondence:} \href{mailto:bbianchi@dc.uba.ar}{bbianchi@dc.uba.ar;}
   \href{mailto:yair.lakretz@gmail.com}{yair.lakretz@gmail.com}
 }
}

\begin{document}
\maketitle
\begin{abstract}
Human readers can accurately count how many letters are in a word (e.g., 7 in ``buffalo''), remove a letter from a given position (e.g., ``bufflo'') or add a new one. The human brain of readers must have therefore learned to disentangle information related to the position of a letter and its identity. Such disentanglement is necessary for the compositional, unbounded, ability of humans to create and parse new strings, with any combination of letters appearing in any positions. Do modern deep neural models also possess this crucial compositional ability? Here, we tested whether neural models that achieve state-of-the-art on disentanglement of features in visual input can also disentangle letter position and letter identity when trained on images of written words. Specifically, we trained beta variational autoencoder ($\beta$-VAE) to reconstruct images of letter strings and evaluated their disentanglement performance using CompOrth - a new benchmark that we created for studying compositional learning and zero-shot generalization in visual models for orthography. The benchmark suggests a set of tests, of increasing complexity, to evaluate the degree of disentanglement between orthographic features of written words in deep neural models. Using CompOrth, we conducted a set of experiments to analyze the generalization ability of these models, in particular, to unseen word length and to unseen combinations of letter identities and letter positions. We found that while models effectively disentangle surface features, such as horizontal and vertical `retinal' locations of words within an image, they dramatically fail to disentangle letter position and letter identity and lack any notion of word length. Together, this study demonstrates the shortcomings of state-of-the-art $\beta$-VAE models compared to humans and proposes a new challenge and a corresponding benchmark to evaluate neural models.
\end{abstract}

\section{Introduction}
\label{sec:intro}
Reading is an invention of human modern culture. Unlike other domains of visual processing such as of faces, reading skills are not innate and require extensive practice \citep{dehaene2015illiterate}. Reading acquisition therefore must rely on the development of a new neural mechanism in the human brain. Given that letters are the building blocks of words, to process new words, the neural mechanism needs to identify single letters in the input, their positions in a word, and compose this information to process entire words. While brain imaging has localized where in the brain visual processing of words occurs \citep{cohen2002language}, what is the precise neural mechanism that enables us to recognize words is largely unknown.

Recent advances in deep neural network models have drastically improved the accuracy on Optical Character Recognition (OCR) tasks \cite{li2023trocr}. Deep neural models can now achieve similar-to-human performance on a variety of tasks, including vision and language. Although neural models are often considered black boxes, full access to their neural computations during processing is possible. Analyzing the properties of these networks provides new opportunities to study neural mechanisms underlying orthographic processing, and in particular, into how letter-identity and letter-position information are extracted from raw images and then composed together to encode whole words.

\begin{figure*}[ht!]
    \centering
    \includegraphics[width=.9\textwidth]{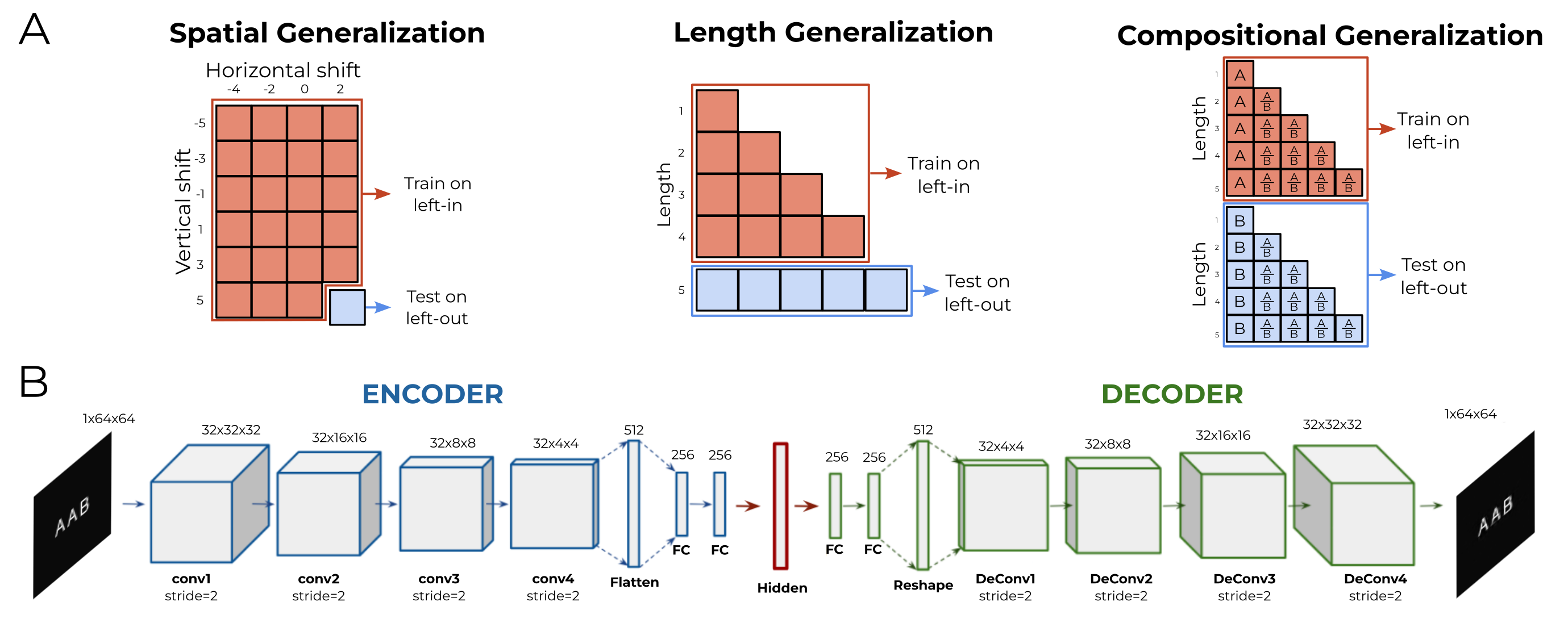}
    \caption{\textbf{(A)} The CompOrth Benchmark: schemes of the three types of generalization tests. \textbf{(B)} An illustration of the architecture of a auto-encoder for processing images of written words.}
    \label{fig:intro}
\end{figure*}

To study this question in neural models, we developed \textit{CompOrth} - a battery of tests, which evaluates compositionality in models and their generalization performance. CompOrth provides several tests, which can be used to both evaluate the `behavioral' performance of the models, as well as study \textit{neural} mechanisms in the model. The tests are designed in a way that directly probes the question of whether a neural model extracts and disentangles letter-identity and letter-position information from raw images, and whether it can compose them together to encode entire words. 

The main hypothesis of this work is that vision models can achieve such \textit{functional} disentanglement of letter identity and position, required to succeed on CompOrth, by \textit{neurally} disentangling this information, representing it in separate units of the model. This is since neural disentanglement could facilitate composition of letter identity and letter position in downstream computations, thus improving generalizations to unseen combinations of letters and positions. Beta Variational Auto-Encoders ($\beta$-VAEs) are a leading model for neural disentanglement \citep{higgins2017betavae}, and, interestingly, they have been shown to also align with neural activity recorded from the primate cortex \citep{higgins2021unsupervised}. Here, we thus study $\beta$-VAEs using CompOrth, testing whether strong neural disentanglement in the models could lead to improved performance on CompOrth, given its requirement for both disentanglement and compositional abilities as of humans.

Our results show that $\beta$-VAEs learn to disentangle surface properties of written words, such as `retinal' horizontal and vertical position. 
However, $beta$-VAEs dramatically fail to disentangle identity and positional information of letters, and to combine them together. We show that such disentanglement of letter identity and position is lacking both at the behavioral and neural levels of the model. Furthermore, we show that $beta$-VAEs lack a robust notion of word length, failing to generalize to unseen word lengths in CompOrth. 
Finally, we provide arguments for why other types of neural models might suffer from the same limitation, and therefore suggest CompOrth as a challenging benchmark for future models.

\section{Related Literature}
\label{sec:leterature}

Literature on cognitive sciences contains competing theories about how words are neurally encoded in the human brain. Based on experiments in humans, early studies theorized the presence of letter combination detectors \cite{dehaene2005neural}, i.e., letters combine to form bigrams, which are further combined to form larger n-grams and finally words. Other theories suggested neural encoding based on open bigrams - neurons that are tuned for higher-order combinations of letters, which are not necessarily adjacent \cite{grainger2004does}. Most recently, a compositional neural encoding scheme was suggested, whereby letter identities and letter positions directly combine to form words \cite{agrawal2020compositional}, in line with early suggestions in the literature \cite{davis2004letter}. 

Modern neural models now provide new opportunities to study how words can be neurally encoded, possibly informing the above debate. One of the computational tasks in the field, most closely related to reading, is Optical Character Recognition (OCR). Various neural models were suggested for OCR, including Recurrent Neural Networks \cite{breuel2013high}, Convolutional Neural Networks (CNNs) \cite{zhang2017cnn}, Transformers \cite{li2023trocr,azadbakht2022multipath}, and hybrid architectures \cite{naseer2019meta,jain2017unconstrained}. Some studies have also tried to identify neural mechanisms in models trained on OCR, or similar tasks. For example, \citet{hannagan2021emergence} studied neural activations of units of CNNs that were trained to recognize images of objects and words. They found that a compositional encoding scheme emerged also in the models. During training, the models developed special units that encode either letter identity or ordinal letter position in a word (but not n-grams). This evidence from neural models thus provide support in favor of a compositional code based on single letters and their positions \cite{agrawal2020compositional}.

In OCR, the goal is typically to identify models that can recover texts from noisy images. This noise can originate from handwritten texts with different writing styles and scripts \cite{baldominos2019survey}, or from photos or scans of deteriorated materials \cite{fontanella2020pattern}. Accordingly, several benchmarks were proposed in the past \cite{shi2016end,lyu2022maskocr,du2022svtr,wang2021two,du2022svtr}, targeting issues of capturing global semantic context \cite{yu2020towards, wan2020textscanner, cui2021representation,bhunia2021joint}, text in different orientations \citet{zhang2020autostr, yan2021primitive}, dependence between the visual processing model and the language model \citet{fang2021read, bautista2022scene}, degraded images quality \cite{mou2020plugnet}, the misrecognition on contextless text images \cite{yue2020robustscanner}, unseen character sequences \cite{bhunia2021towards} and datasets with few labels \cite{baek2021if}.

However, in contrast, the primary objective of CompOrth is to identify models that achieve \textit{compositionality}. That is, models that can recognize words with new combinations of letter identities and letter positions, unseen during training, similarly to how humans process new words. To achieve compositionality, a model would need to learn to identify single letters and then efficiently compose them with their positions, recursively, when encoding entire words. To focus on compositionality, CompOrth thus simplifies much of the problem by eliminating several sources of noise present in OCR corpora, as described below.

\section{General Setup}
\subsection{The CompOrth Benchmark}
\label{sec:stimuli}

\paragraph{Stimuli:}
The stimuli generated for CompOrth were designed to probe the encoding of single letters and their positions, and therefore they minimize the amount of other types of information the model needs to learn. For this, the strings (hereafter, `words') in each test contain only two letters (e.g., ``A'' and ``B''), in the same case (upper case), the same font (`Arial') and the same letter size (`12'). For each combination, we generated all 62 possible words of 1 to 5 letters (e.g., ``A'', ``B'', ``AA'', ``AB'', ..., ``BBBBA'', ``BBBBB''). Also, for each word, images were generated by varying their location in the image (hereafter, `retinal' location), and the spacing between characters, resulting in a total of 11,904 images. All images were generated with white letters on a black background (Figure \ref{fig:supp-stimuli}).

The retinal location was varied by shifting the position of the string from the center of the image both vertically and horizontally. Specifically, words with zero displacement in both axes have their center at the center of the image. Meanwhile, words with a displacement of 1 in both axes will have their center shifted one pixel up and one pixel to the right. On the horizontal axis, the strings were moved from 4 pixels to the left to 4 pixels to the right. On the vertical axis, the strings were moved from 4 pixels up to 4 pixels down (Figure \ref{fig:supp-stimuli}). 

Spacing variation was introduced to decouple retinal (absolute) location of a letter within an image and its (relative) position within a word. This factor modifies the default font spacing and moves each character a certain number of pixels to the left or right. Ensuring that contiguous characters do not overlap, the spacing was varied from 2 pixels to the left to 2 pixels to the right, with 0 being the default spacing.

\paragraph{Three Generalization Tests:}
\label{sec:tests}

CompOrth contains three groups of tests, of increasing complexity: (1) \textit{Spatial Generalization} (Figure \ref{fig:intro}A-Left), which evaluates generalization across `retinal' positions, (2) \textit{Length Generalization} (middle), which evaluates generalization to unseen word lengths (shorter or longer), and (3) \textit{Compositional Generalization} (right), which evaluates generalization to unseen combinations of letter identities and letter positions, thus evaluating whether the models develop abstract notion of letter position and letter identity. Each test consists of several splits into training and test sets, ensuring that in each split, the test set contains images generated from different combinations of factors than those in the training set. That is, each of the sets was generated by selecting one level of a given generative factor (e.g., bottom right-most shift of retinal position, as in Figure \ref{fig:intro}A-Left) and removing it from the training set. The corresponding test set then comprises all the images generated with the left-out level of the factor (blue square). Specifically,

\noindent\textbf{Spatial Generalization:} For each possible combination of x-y-shift, all words with that combination are used as the left-out while the rest of the words are used as the left-in set (Figure \ref{fig:intro}A, left).

\noindent\textbf{Length Generalization:} For each length, all words with that length are used as the left-out set, while the rest of the words are used as the left-in set (Figure \ref{fig:intro}A, middle). 

\noindent\textbf{Compositional Generalization:} For each relative position of each letter (e.g., `A' in 2$^{nd}$ position), all words with that combination (e.g., ``AA'', ``BA'', ``AAA'', ``BAB'', etc.) are used as the left-out set, while the rest of the words are used as the left-in set (Figure \ref{fig:intro}A, right).

\subsection{Models}

\subsubsection{Model Architecture}
\label{sec:model-arch}

We benchmarked CompOrth with Variational Auto-Encoders (VAEs; \citealp{kingma2013auto}), including a more recent variant of this model, known as $\beta$-VAE \citep{higgins2017betavae}. Following \citet{higgins2017betavae}, the models comprised of a 4 convolutional and 2 fully connected layers encoder. The decoder has a mirrored architecture (Figure \ref{fig:intro}B). 
For CompOrth, an advantage of evaluating auto-encoders is that they can be tested on unseen words. In contrast, standard feed-forward classifiers have a finite set of output units \cite{hannagan2021emergence}, corresponding to different words, and therefore evaluating the model on unseen words is often not possible without retraining the model. Moreover, $\beta$-VAEs can be optimized to achieve neural disentanglement, which encourages the activity of single units in the latent layer to encode different generative factors of the training data, such as letter identity and letter position.

\subsubsection{Model Training}
\label{sec:model-train}

For training, we used a batch size of 64 samples. Training and Evaluation were conducted with Nvidia Quadro RTX 8000 48 GB GPUs. The whole experiment, including grid search, took about 72 hours. For the grid search, we optimized for the following hyper-parameters: Beta ($2^i, i \in \{0, ..., 7\}$), size of the latent layer ($2^i, i \in \{3, ..., 7\}$) and initial learning rate ($10^i, i \in \{-4, ..., -2\}$). The optimal learning rate across all combinations in the grid search was consistently $10^{-4}$. We set the maximal number of epochs to 1000, which we verified to be large enough to reach full convergence in all cases.

\begin{figure*}[t!]
    \centering
    \includegraphics[width=\textwidth]{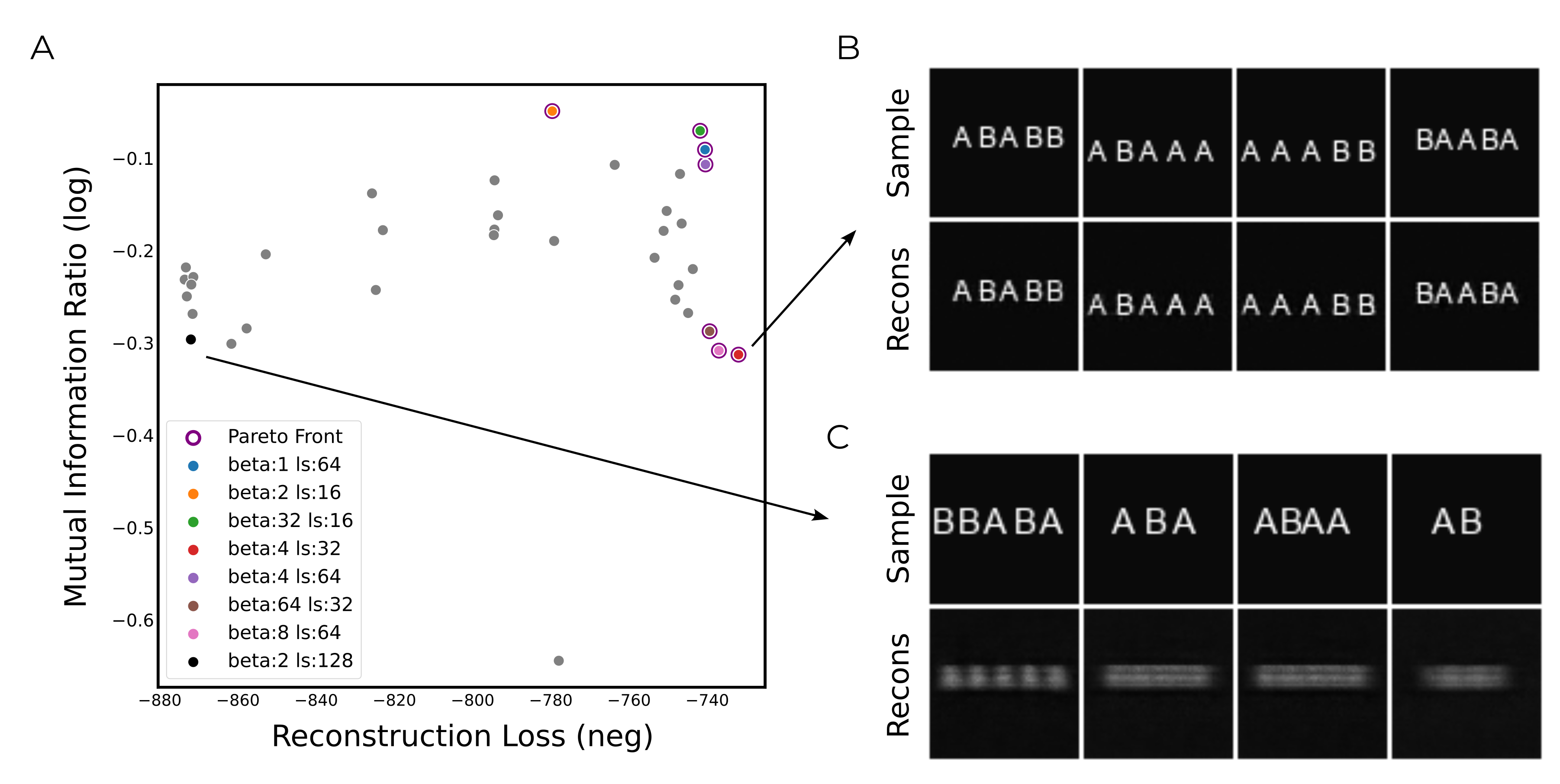}
    \caption{\textbf{Model Selection} \textbf{(A)} Reconstruction loss against Mutual Information Ratio. Each dot represents a model. Orange dots represents models from which samples were taken as examples in the right panels. \textbf{(B)} Reconstruction examples from the model with best reconstruction loss and MIG ($\beta$=4, Latent Size=32, Learning Rate=0.0001). \textbf{(C)} Reconstruction examples from a model with relatively poor reconstruction loss ($\beta$=2, Latent Size=128, Learning Rate=0.0001). Models marked with purple circles represent the Pareto Front.}
    \label{fig:selection}
\end{figure*}

\subsubsection{Model Evaluation}
\label{sec:model-eval}

\paragraph{Reconstruction Loss} For model selection, we used the standard reconstruction loss for visual AEs, which is calculated based on the pixel-by-pixel difference between the original and the reconstructed images.

\paragraph{Reconstruction Accuracy} However, for testing compositionality with CompOrth, mere reconstruction loss might be little informative. This is since a large number of pixels in the images could be simply black, therefore the reconstruction loss might be low albeit poor recognizability of the reconstructed word. Moreover, the reconstruction of blurry unidentifiable letters can help to lower reconstruction loss without genuinely improving the word recognizability (see examples in Figure \ref{fig:selection}B). This is crucial when evaluating the model on unseen images when the model is faced with the compositionality challenge. To address this, we defined another evaluation metric -- \textit{Reconstruction Accuracy}, which quantifies word \textit{recognizability}. For this, we trained another model - a standard feed-forward CNN-based classifier, using the entire set of words in CompOrth, and used it to evaluate the reconstruction quality of the $\beta$-VAE. We refer to this model as the \textit{Evaluator model}. The Evaluator was trained with the original images only, until it reached perfect performance, and presented, at test time, with reconstructed images from the $\beta$-VAEs. The Evaluator contained 2 consecutive convolutional layers and 2 linear layers, and the last layer was a softmax layer across the 62 possible words. Reconstruction accuracy was then defined as the output probability of the CNN classifier for the desired word. For example, given an image with an unseen image containing ``ABABA'', the reconstructed image (the output of the VAE) was presented as an input to the CNN classifier. The output probability of the CNN classifier, which corresponds to the target word ``ABABA'', was then considered as the reconstruction accuracy of the model for this image.

\paragraph{Metrics for Disentanglement (MIG and MIR)}
To quantify neural disentanglement, previous work \cite{chen2018isolating} used the Mutual Information Gap (MIG). However, \citet{whittington2022disentanglement} suggested an alternative measure called the Mutual Information Ratio (MIR). The main difference between these two metrics is that the former scores high if each factor is encoded in a single neuron. In contrast, the latter scores high when each neuron responds to a single factor. Thus, the second one allows high score also when multiple neurons respond to a single factor, as long as the response is to a unique factor. Since compositionality can be achieved even if several neurons disentangle and extract the same type of information, we focused on MIR. We therefore used MIR to test the hypothesis that neural disentanglement may facilitate behavioral disentanglement of letter-identity and position information required for CompOrth.

\section{Results}
\label{sec:results}

\subsection{Model selection}

We first optimized for the hyperparameters of the $\beta$-VAE models, using nested cross-validation. Figure \ref{fig:selection}A illustrates model selection, by showing both the reconstruction loss and the MIR for all models in the grid-search. Since both good reconstruction and disentanglement are desired properties of a model, the optimal models lie on the Pareto front (purple circles) of the problem. No other models outperform them in both criteria simultaneously. In what follows, we therefore report results based on average performance across all optimal models on the Pareto front. We later analyze particular cases from these models (section \ref{sec:perturbation}).

To illustrate reconstruction ability of the models, Figure \ref{fig:selection}B\&C show examples from two models - one from the Pareto front, having good reconstruction performance (with $\beta=4$, \textit{latent-layer size} $=32$, \textit{learning rate} $=10^{-3}$) and the other with a low one ($\beta=4$, \textit{latent-layer size} $=128$, \textit{learning rate} $=10^{-5}$). In both panels, the top row shows examples from the original images, and the bottom row shows their corresponding reconstructions. In the case of low reconstruction performance, the reconstructed images are blurry, with only 'retinal' location preserved.

\begin{figure*}[h!]
    \centering
    \includegraphics[width=.7\textwidth]{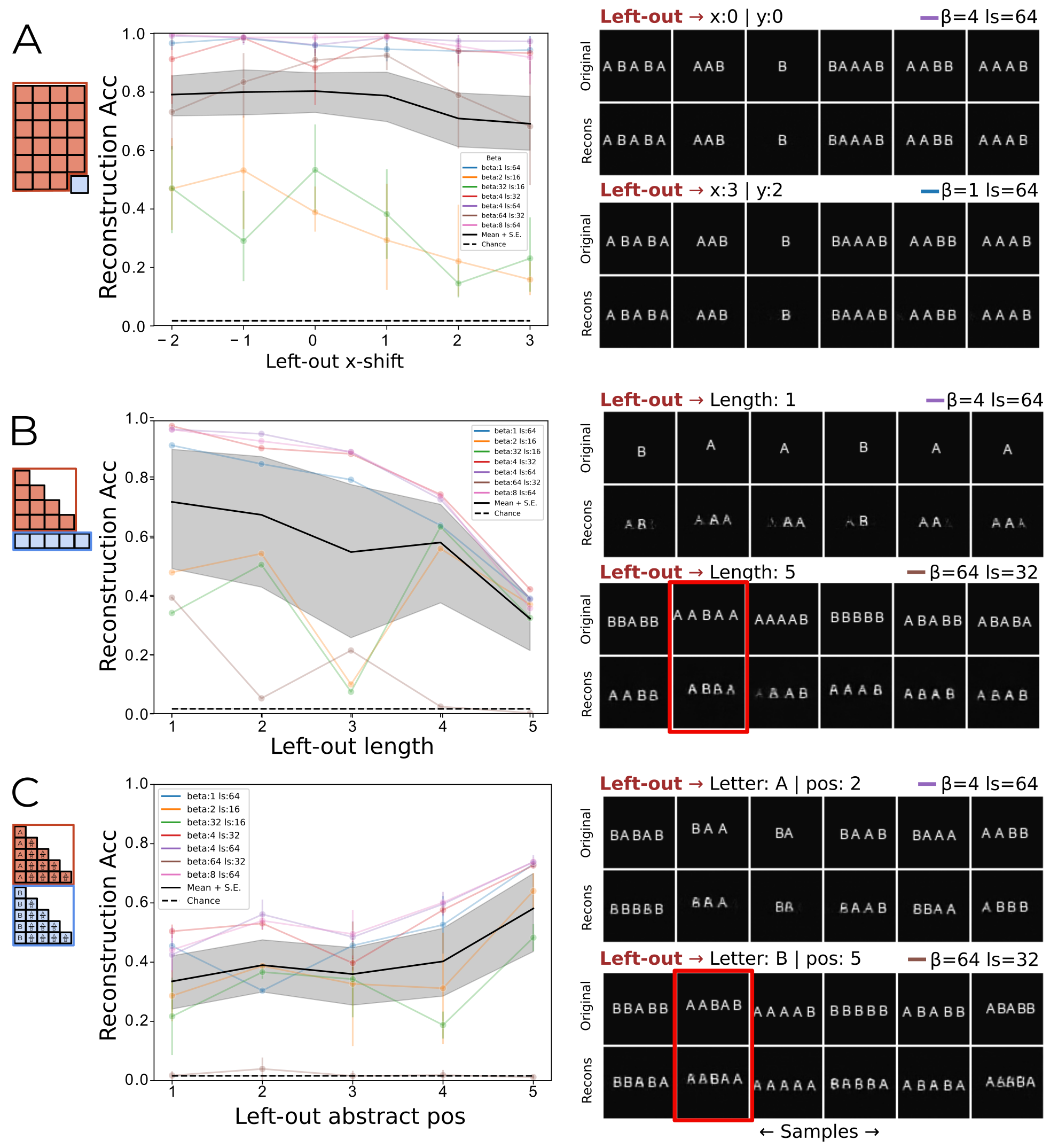}
    \caption{\textbf{Results on CompOrth tests.} Average reconstruction loss on the \textbf{(A)} Retinal-Position Test, \textbf{(B)} Word-Length Test, and \textbf{(C)} the Abstract-Position Test. Dashed lines mark the chance level for the classifier ($chance=1/62$). On the right of each panel, several examples are shown for how the model reconstruct test images. Note the red marking on the images, which highlight the type of errors the model makes. }
    \label{fig:gen}
\end{figure*}

\subsection{Behavioral Evaluation using CompOrth}
\paragraph{Spatial Generalization -- $\beta$-VAEs can generalize to unseen `retinal' locations:}
Figure \ref{fig:gen}A-Left shows the mean generalization performance to unseen retinal positions, for all models from the Pareto front (mean performance in black). On average, the models show good ability to reconstruct words in positions where they were not seen during training. Except for three models on the Pareto front ($layer-size (ls)=16$, orange and green lines, and $beta=64$, brown line) all other individual models achieve mean accuracy above $90\%$ for all spatial generalizations, also for vertical generalization (Figure \ref{fig:supp-retinal_pos_byBeta}) . Figure \ref{fig:gen}A-Right further shows reconstruction examples: each of the plots shows 6 random samples (top) and their reconstruction (bottom) by a given model. Overall, the reconstruction is, qualitatively, similar to the original image, even for models with relatively low performance.

\paragraph{Length Generalization -- $\beta$-VAEs fail to generalize to longer word lengths:} Figure \ref{fig:gen}B-Left shows the reconstruction accuracy for all left-out word lengths, as measured with the Evaluator Model (section \ref{sec:model-eval}). Overall, reconstruction accuracy for short left-out word length is high, in particular for the four models with also better performance on spatial generalization. However, a qualitative inspection of random samples from these models (\ref{fig:gen}B-Right, for examples) show that letter parts are nonetheless present in the reconstruction of images with a single letter, remnants from the longer words in the training data. 
In general, for longer word lengths, generalization performance decreases. In particular, generalization performance for words with five letters is lowest. A qualitative inspection shows that, indeed, the reconstruction of words with five letters contains in many cases only four letters (red rectangle).

\paragraph{Compositional Generalization: $\beta$-VAEs fail to generalize to unseen compositions of letter identity and letter position:} Finally, the compositional test assessed the ability of the models to develop an abstract understanding of letter position. Overall, results show that the compositional test was most challenging for the models compared to the other two tests (Figure \ref{fig:gen}C-Left), with one model's performance approaching chance level. Figure \ref{fig:gen}C-Right illustrates the type of errors the models make (e.g., in red rectangles). For example, when presented with strings where `B' was never present in the 5$^{th}$ position in the training data, the models `hallucinate' an `A' in this position. Similar errors occur when `B' or `A' were omitted from other positions during training.

\subsection{Neural Evaluation using Perturbation Experiments}\label{sec:perturbation}

\begin{figure}[t]
    \centering
    \includegraphics[width=\columnwidth]{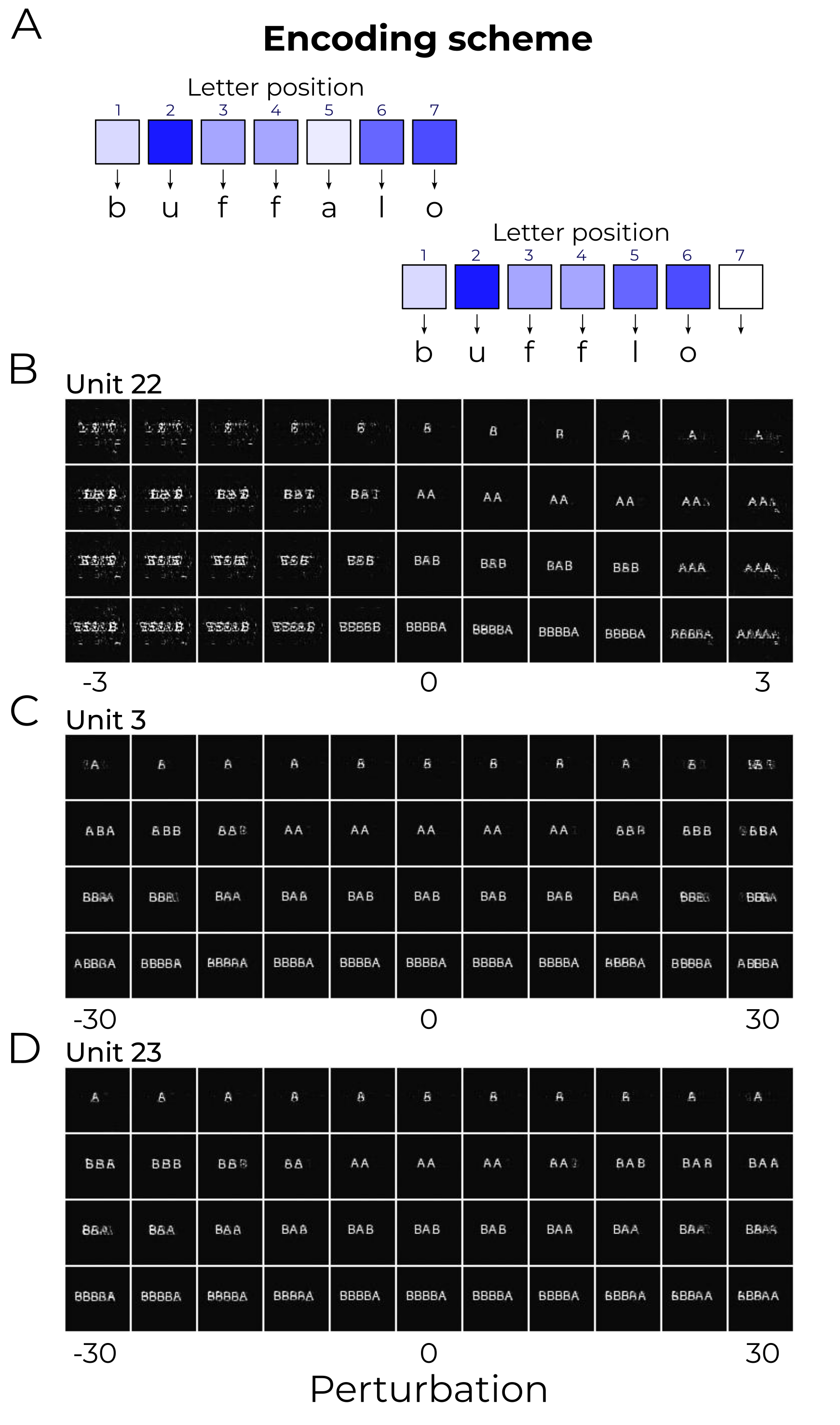}
    \caption{\textbf{Neural Perturbation Analyses} \textbf{(A)} Examples of an hypothetical encoding scheme: single neurons for positions, where the degree of activation indicates which letter is present in that position. \textbf{(B-D)} Perturbation results for example units from a model with strong performance on CompOrth ($\beta=4$, \textit{latent-size}$=32$). Each row represents different samples (word images), columns represents different levels of perturbation.}
    \label{fig:traversals}
\end{figure}

We next studied to what extent beta-VAEs develop neural disentanglement of letter position and letter identity information. Figure \ref{fig:traversals}A illustrates what such neural disentanglement could look like -- it shows a possible encoding scheme, where different units of the latent layer encode for different positions in the word, and different levels of activity encode for different level identities. Therefore letter identity is independently encoded of letter position.

To test whether such neural disentanglement emerges in the model, even partly, we conducted a perturbation experiment. That is, given an input image from the training set, we computed the neural activations at the latent layer of the model, and then separately for each unit, we systematically perturbed its activity to different levels. After each perturbation, we then reconstructed the image. The difference between the input image and the reconstructed image (after perturbation) is revealing about the information that the perturbed unit encodes. For example, if a model developed neural disentanglement of letter position and letter identity (Figure \ref{fig:traversals}A), then perturbing one of the latent units can cause a replacement of one letter with another one, in only the perturbed position.

Figures \ref{fig:traversals}B-D show examples from the perturbation experiments, from the model with the best reconstruction loss and strong performance on CompOrth ($\beta=4$, \textit{latent-size}$=32$; red lines in Figure \ref{fig:gen}). Each panel corresponds to a latent unit, rows of each panel correspond to different samples (input images) and columns to different levels of perturbation.

Unit 22 in the latent layer of the model (Panel B) illustrates spatial encoding -- positive perturbations caused increased translation of the reconstructed string along the vertical axis. However, spatial information is not fully disentangled from other factors, as perturbations also led to changes in the letters of the string.  Unit 3 of the model (Panel C) illustrates word-length encoding -- in some cases, increased perturbations added letters to the strings. However, here too, other types of information, such as letter identity, seem to be encoded in this unit. Finally, unit 23 (Panel D) illustrates the encoding of letter identity -- increased perturbations led, in some cases, to change of letter identity. Figures \ref{fig:supp-traversals-complete1} \& \ref{fig:supp-traversals-complete2} show perturbation effects for all 32 units of the model.

However, analyzing all 32 latent units in the model, we did not identify an encoding scheme that fully disentangle letter identity and letter position (e.g., Figure \ref{fig:traversals}A). The examples above provide only sporadic evidence in this direction, from example units, and none of the models on the Pareto front developed strong disentanglement. This is, in fact, consistent with the relatively poor performance of all models on the Compositional-Generalization test in CompOrth. Limited neural disentanglement is consistent with poor compositionality and thus low performance on CompOrth.

\subsection{The Relationship between Neural Disentanglement and Compositionality}

While we haven't discovered strong neural disentanglement of identity and position in the previous section, a weak neural disentanglement might have nonetheless emerged in some of the models, which is hard to detect with mere perturbation experiments. Such weak disentanglement would possibly lead to a small, yet significant, improvement in performance on the Compositional-Generalization test in CompOrth. 

We therefore next tested the hypothesis that neural disentanglement facilitates the separation of letter-positions and letter-identity information, and therefore, in turn, their composition. This predicts that models that achieve high neural disentanglement, as measured by MIR (section \ref{sec:model-eval}), will achieve better performance on compositional generalization, as measured by the Compositional-Generalization test in CompOrth. To test this, we computed the correlation between the MIR and reconstruction accuracy on CompOrth for all models on the Pareto front. We found a weak correlation $\rho=0.13$ (Figure \ref{fig:supp-MIR-ACC-Gen}), however, which was not statistically significant ($p-val=0.26$).

\section{Summary and Discussion}
\label{sec:discussion}

We introduced CompOrth, a novel benchmark for evaluating orthographic processing in visual models. The primary goal of CompOrth is to assess compositionality -— the ability of a model to generalize to new combinations of letter identities and positions beyond the training set. This task is considered trivial for humans, so passing the CompOrth test is essential for a model to be regarded as achieving human-like performance. 

We tested a large number of variational auto-encoders (VAEs), including $\beta$-VAEs, a variant that encourages the emergence of neural disentanglement in VAEs. We found that all models failed on the target compositional tests of CompOrth, failing both to generalize to new combinations of letters and positions as well to longer word lengths. This shows that the models did not capture the underlying process that generated the data, whereby images were sampled from combinations of letters in varying positions. Instead, the results suggest that the models rely on \textit{memorizing} the training data. This is observed in examples where the model 'hallucinates' a letter in a given position, or adding a new letter, failing to reconstruct the input image. 

Theoretically, the information bottleneck principle in auto-encoders could encourage models to learn underlying patterns in the data by forming abstract representations and relying on rule-based encoding rather than memorization. However, there's a trade-off with reconstruction accuracy. An excessively constrained bottleneck can compromise reconstruction performance, as illustrated in the hyperparameter optimization process (Figure \ref{fig:selection}).

We further hypothesized that neural disentanglement, which tends to emerge in $\beta$-VAEs with high values of $\beta$, would facilitate the separation of letter-positions and letter-identity information, and in turn, their composition. To test this, we conducted perturbation experiments with $\beta$-VAEs, to see whether some of the units disentangle identity and position information. Exploring all units in the model, we have not identified any such units, which is consistent with the failure of the models on CompOrth. However, a weak neural disentanglement of identity and position may have emerged in some of the models, unobserved by our perturbation experiments. We therefore tested whether there exists a correlation between MIR and CompOrth performance, across all our VAE models. We found no significant correlation between MIR and reconstruction accuracy on CompOrth. 

The observed failure of the models in this study is one more example of the shortcoming of artificial neural networks to dynamically and flexibly bind information, which might be distributedly encode in the network \cite{greff2020binding}, even when neural disentanglement is explicitly optimized, as in $\beta$-VAEs. This binding problem affects the capacity of the models to achieve compositional ability by manipulating symbols (letters) and combine them in various, unbounded, ways, as humans \citep{fodor1988connectionism}. Similar limitation, for similar reasons, was observed also in language models \citep{deletang2022neural}. However, unlike existing benchmarks for testing compositionality in language models (e.g., \citealp{lake2018generalization}), for orthographic processing and OCR, there are no existing targeted benchmarks. CompOrth therefore aims to fill in this gap, by providing means to evaluate compositionality in vision models. Our study shows that $\beta$-VAE models do not achieve good compositionality and achieve only a limited neural disentanglement of letter position and identity, which suggests CompOrth as a simple yet challenging test for future models.

\section*{Ethical statement}
This paper presents work whose goal is to bridge closer the fields of Machine Learning and Psycholinguistics; being theoretical in nature, we believe that no societal risks need to be specifically highlighted.

\section*{Limitations}
This study investigates the ability of a neural architecture to disentangle relevant information in the input for compositional generalization. One possible limitation is that the models explored here were trained on the CompOrtho dataset only. However, we note that while the models were trained from scratch on CompOrth, the challenges posed by the benchmark are also applicable to pre-trained models. These models can be refined and evaluated using the same approach (Figure \ref{fig:intro}A) to assess their compositional generalization capabilities.

\bibliography{latex/custom}

\newpage
\appendix

\renewcommand{\thefigure}{A.\arabic{figure}}
\setcounter{figure}{0}

%


\begin{figure*}[ht]
    \centering
    \includegraphics[width=.7\textwidth]{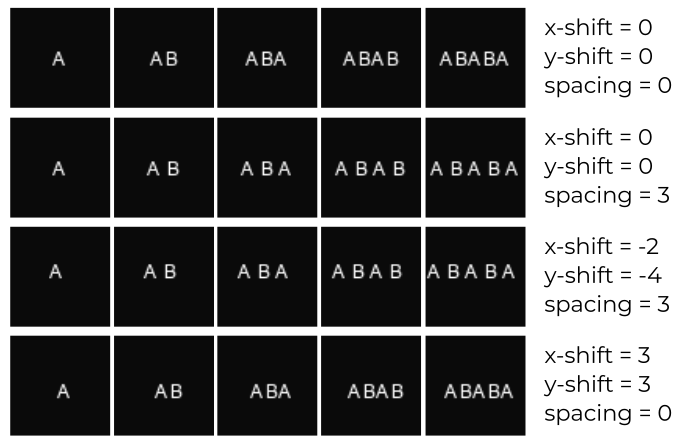}
    \caption{\textbf{Examples from the generated dataset.} All the images are comprised by a string of 1 to 5 letters, using only the uppercase characters A and B. To generate variations of this strings the spacing and the x and y position were modified.}
    \label{fig:supp-stimuli}
\end{figure*}

\begin{figure*}[ht]
    \centering
    \includegraphics[width=\textwidth]{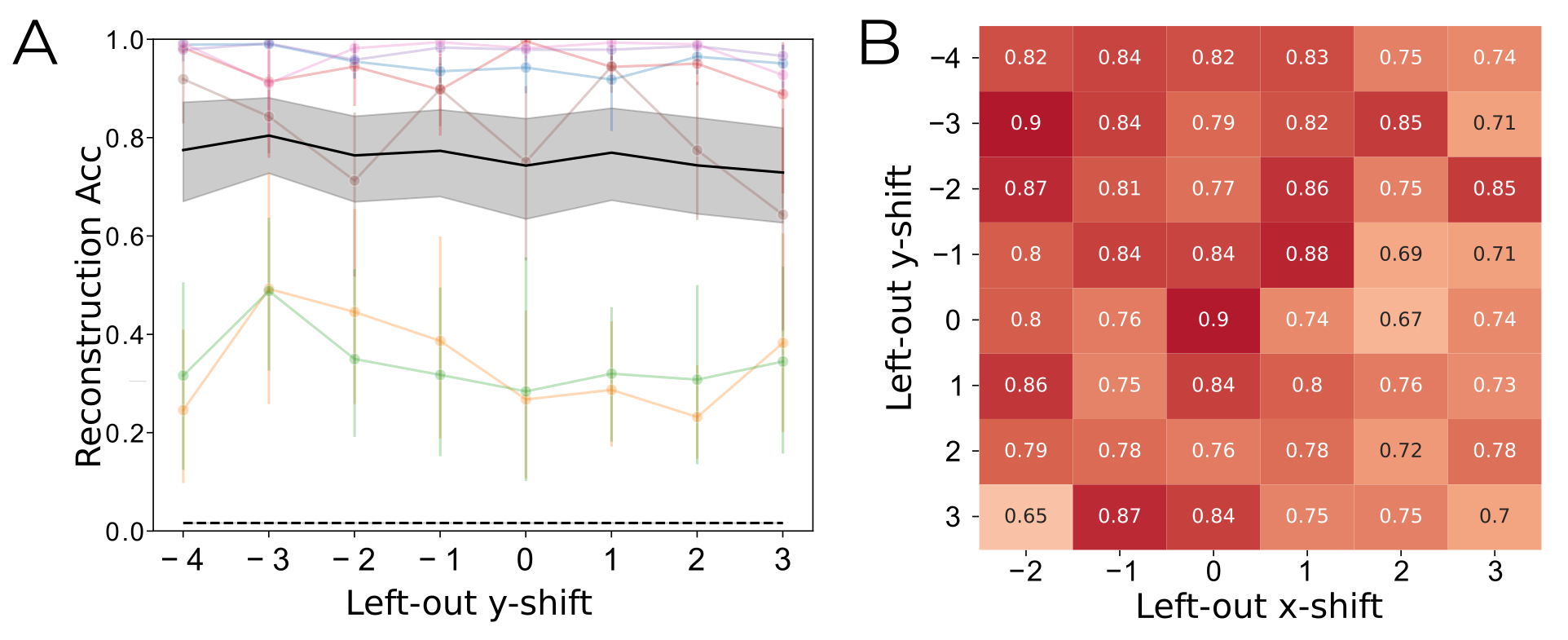}
    \caption{\textbf{Reconstruction accuracy for Spatial Generalization test. (A)} averaged by model, across y-shift; and \textbf{(B)} average for each combination of x- and y-shift, across models.}
    \label{fig:supp-retinal_pos_byBeta}
\end{figure*}

\begin{figure*}[ht]
    \centering
    \includegraphics[width=\textwidth]{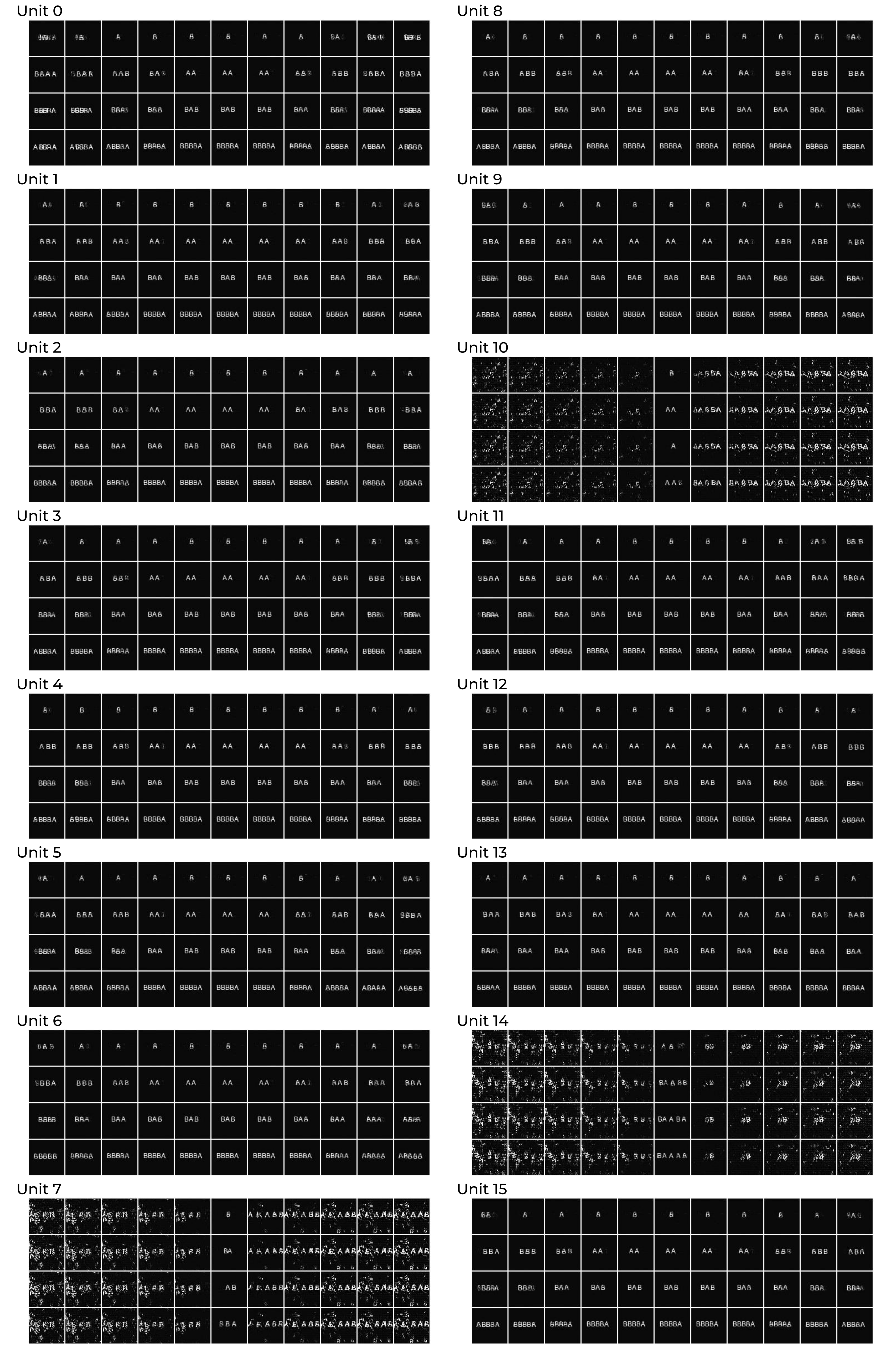}
    \caption{\textbf{Neural Perturbation Analyses.}Perturbation results for example units from a model with strong performance on CompOrth ($\beta = 4$, $latent-size= 32$). First 16 neurons of the model.}
    \label{fig:supp-traversals-complete1}
\end{figure*}

\begin{figure*}[ht]
    \centering
    \includegraphics[width=\textwidth]{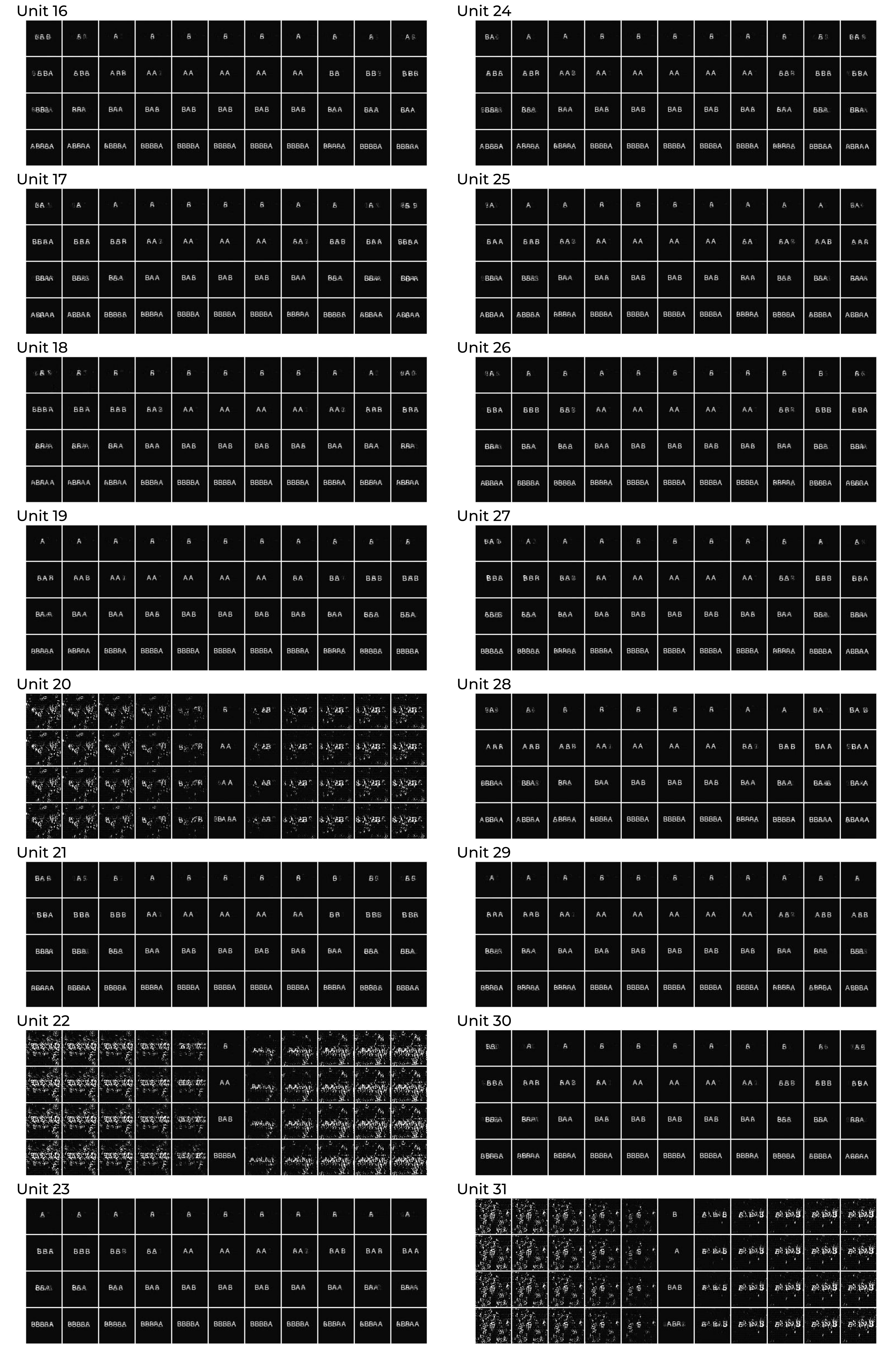}
    \caption{\textbf{Neural Perturbation Analyses.}Perturbation results for example units from a model with strong performance on CompOrth ($\beta = 4$, $latent-size= 32$). Last 16 neurons of the model.}
    \label{fig:supp-traversals-complete2}
\end{figure*}

\begin{figure*}[ht]
    \includegraphics[width=.9\textwidth]{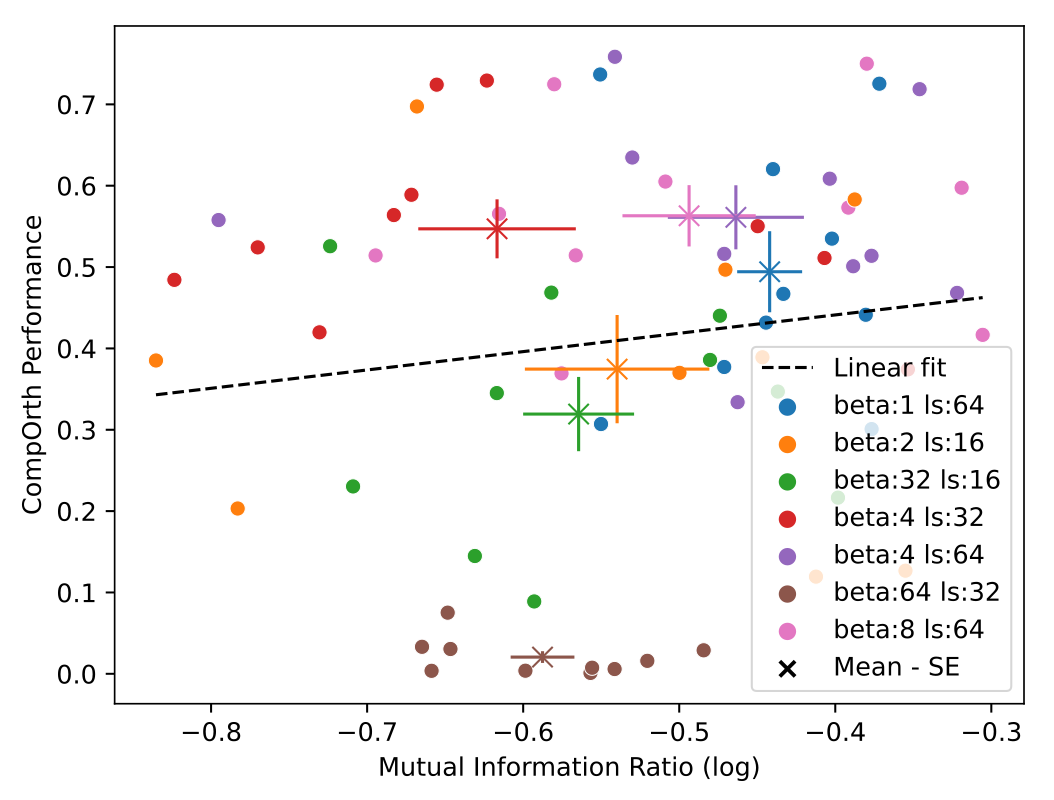}
    \centering
    \caption{\textbf{Mutual Informtion Ratio (MIR) vs Reconstruction Accuracy for Compositional Generalization analyses.} MIR was calculated using the word identity Encoding Scheme of Figure \ref{fig:traversals}A for 5-letter words over all the models in the Pareto Front. Each model is encoded with one color. Each circle of a given model corresponds to the results of that model in a test split of the Compositional Generalization test. Crosses represents the mean for each model, across all the test splits. Error bars corresponds to the Standard Error of the mean. Dashed line represents the lineal fit of the data. The Pearson correlation resulted in $\rho=0.12$ and $p-val=0.30$.}
    \label{fig:supp-MIR-ACC-Gen}
\end{figure*}
\end{document}